\documentclass[letterpaper]{article}
\usepackage{aaai2027}
\usepackage[hyphens]{url}
\usepackage{graphicx}
\usepackage{natbib}
\usepackage{caption}
\usepackage{booktabs}
\title{Explaining Basketball Shot Classifications with Grad-CAM: An Investigation of Shortcut Reliance with Spatiotemporal Correlations}
\author{
    Hongyi Tian, \textsuperscript{1}
    Robin Cohen \textsuperscript{2}
}
\affiliations{
    Computer Science, University of Waterloo\\
    200 University Ave W, Waterloo, ON, N2L 3G1, Canada\\
    +14374366580; hongyihenrytian@gmail.com\textsuperscript{1}; rcohen@uwaterloo.ca\textsuperscript{2}\\
}

\begin{document}
\maketitle

\begin{abstract}

% Basketball action recognition is used in high-stakes scenarios to inform broadcast annotation, specialized coaching, and scouting tools, yet the evidence used by its deep-learning models is poorly understood. This paper investigates whether an R(2+1)D-18 model trained on the Basketball-51 dataset relies on spurious spatiotemporal correlations when classifying basketball shot outcomes and types. To audit shortcut learning, Gradient-weighted Class Activation Mapping (Grad-CAM) generated spatiotemporal saliency maps, whose faithfulness was evaluated through progressive perturbation. Twenty saliency maps were then visually reviewed for recurring salient regions. Grad-CAM highlighted spatially central regions of clips including the shooter, net, and occasionally the courtside crowd. Blurring interventions were conducted to support these patterns. Central blurring resulted in an accuracy of 43.7\%, compared to the original test accuracy of 69.3\% and random-blurred accuracy of 50.9\%. The additional 7.2 percentage-point reduction from central blurring compared to random blurring suggests reliance on task-relevant visual regions and highlights the need for continual audits to ensure trustworthy, equitable sports analytics.

Basketball action recognition is used in high-stakes scenarios to inform such tasks as specialized coaching, yet the evidence used by its deep-learning models is poorly understood. This paper investigates whether there is evidence that an R(2+1)D-18 model trained on the Basketball-51 dataset relies on spurious spatiotemporal correlations when classifying basketball shot outcomes. Gradient-weighted Class Activation Mapping (Grad-CAM) was used to generate spatiotemporal saliency maps, whose faithfulness was evaluated by progressively blurring the most salient pixels. Twenty saliency maps were then visually reviewed for recurring salient regions. Grad-CAM highlighted spatially central regions of clips including the shooter, net, and occasionally the courtside crowd. Blurring interventions were conducted to support these patterns. Blurring the central region resulted in an accuracy of 43.7\%, compared to the original test accuracy of 69.3\% and random-blurred accuracy of 50.9\%. The additional 7.2 percentage-point reduction from central blurring compared to random blurring suggests that the model relies on task-relevant visual regions;  this highlights the need for continual audits to ensure trustworthy, equitable sports analytics.

\end{abstract}

\section{Motivation}

The black-box nature of shot evaluation deep-learning models prevents stakeholders from understanding the model's decisions, potentially leading to incorrect conclusions and mistrust. Explainable artificial intelligence (XAI) techniques can reveal the evidence behind model decisions. For example, Gradient-weighted Class Activation Mapping (Grad-CAM) produces class-discriminative localization maps \citep{selvaraju2017gradcam} to highlight the regions of a video most salient to a model's decision, such as a shooter's body.

One shortcoming that XAI can reveal is shortcut learning, which occurs when a model achieves strong in-distribution performance by exploiting unintended features such as scoreboard changes rather than intended, task-relevant evidence such as shooting posture and net movement \citep{geirhos2020shortcut}. Across arenas and leagues where scoreboard graphics and crowd sizes change, shortcut learning may impair model performance in real-life applications. In coaching and scouting pipelines, such models risk unfair and inaccurate evaluation of athletes, especially across leagues and arenas. Although shot evaluation is an active area of research, the application of XAI in it remains under-analyzed; Caruso et al.~\cite[p.~8]{caruso2025advancements} state that ``no dedicated studies specifically address explainability in basketball action recognition.'' This study investigates whether there is evidence that a 3D-CNN relies on spurious spatiotemporal correlations instead of authentic mechanics to classify shot outcomes.

Bridging action recognition, XAI, and sports analytics is essential to this research. Action recognition provides predictions, XAI produces localized spatiotemporal hypotheses about the evidence used, and sports analytics distinguishes meaningful, task-relevant cues from spurious correlations. Combined, they enable an explanation workflow applicable to sports action models.

\section{Data and Model}

We use the Basketball-51 dataset introduced by Shakya et al.~\cite{shakya2021basketball51}. It contains shooting clips from 51 broadcast NBA games, divided into 8 classes. Each clip retains all visual features of the original broadcast such as on-screen scoreboards and a moving camera. 7,750 clips from 39 games, 76.5\% of games, formed the training dataset, 1,440 clips from 7 games (13.7\%) used in the validation dataset, and 1,121 clips from 5 games (9.8\%) reserved for testing and explanations. The splits contain disjoint games. 

As the baseline model, we fine-tuned a Kinetics-400-pretrained R(2+1)D-18, which factorizes each 3D spatiotemporal convolution into a 2D spatial convolution followed by a 1D temporal convolution \citep{tran2018closer}. Input video clips were resized to $112 \times 112$ pixels across 16 frames. After adaptive global average pooling, the replacement classification head applies a linear transformation from 512 features to eight class scores. All model parameters were fine-tuned using a batch size of 32, a learning rate of $2\times10^{-4}$, and a weight decay of $1\times10^{-4}$. The checkpoint with the highest validation accuracy was selected for subsequent analysis. The model was trained until stopping validation accuracy failed to improve for 2 epochs in a row.

\section{Explanation and Intervention Protocol}

Grad-CAM faithfulness was evaluated on 1121 testing clips. At each subsequent step, an additional 10\% of the highest-ranked spatiotemporal locations was replaced by values from a $7\times7$ Gaussian-blurred version of the corresponding frame. As a control, the same procedure was repeated using a random ordering of spatiotemporal locations. Probabilities were averaged across the 1121 clips at every perturbation level.

Saliency maps of twenty test set clips were visually reviewed for recurring salient spatial regions to create a new dataset by applying Gaussian blur with kernel size 7 on that region in the original test set. A control set is created by blurring a random section of equal size in each test clip. The model's change in accuracy in both new datasets helps determine if shortcut learning is occurring.

\section{Results and Limitations}

Training stopped after epoch 6 as validation accuracy failed to improve for two consecutive epochs. Epoch 4's checkpoint was retained because it achieved the peak validation accuracy of 64.4\%, with a validation
cross-entropy loss of 1.267. 

\begin{figure}[htbp]
    \centering
    \includegraphics[width=0.80\linewidth]{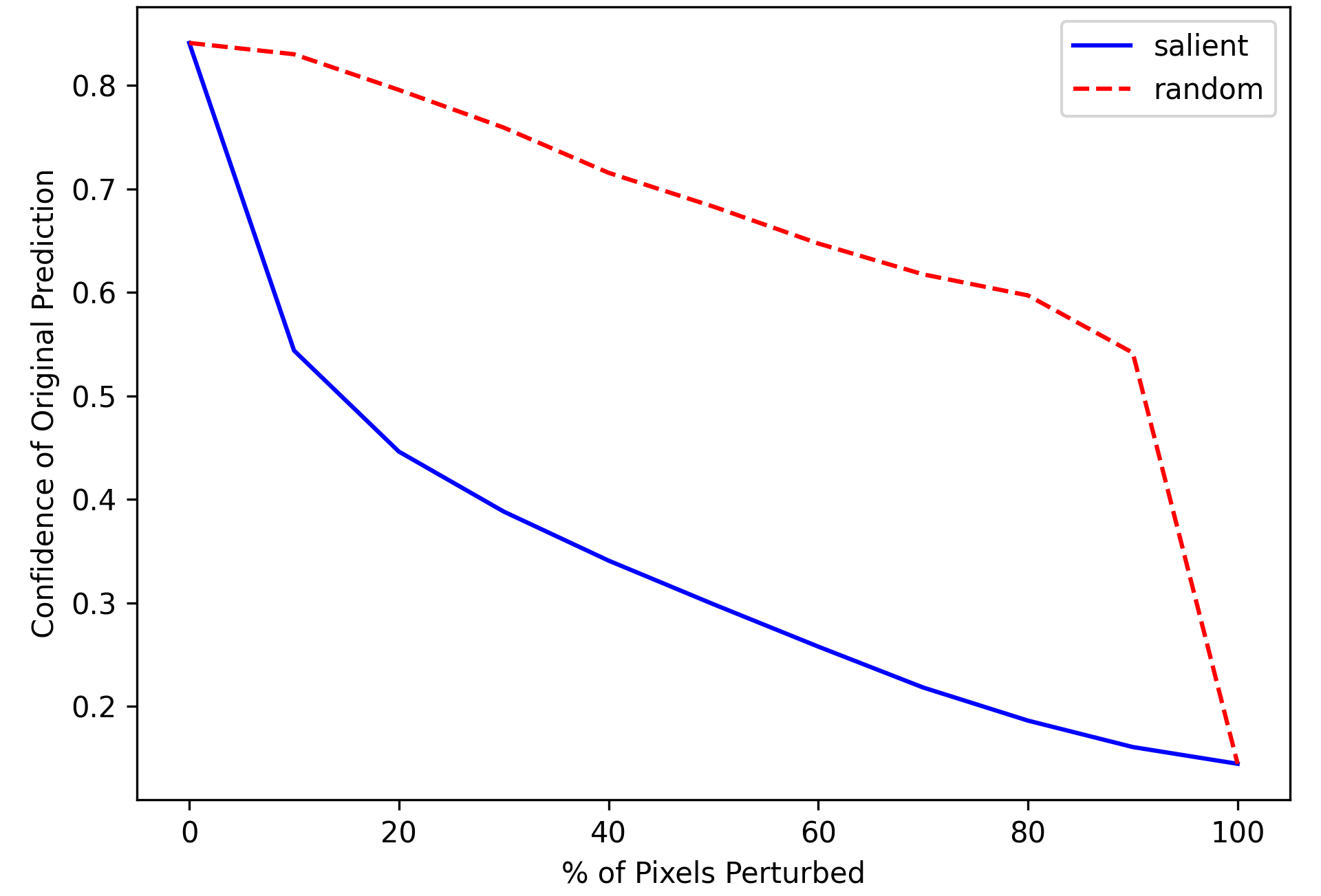}
    \caption{Average confidence of the model's original prediction after perturbations.}
    \label{fig:perturbation}
\end{figure}

Figure~\ref{fig:perturbation} shows the model's softmax probability for its originally predicted label across the perturbation experiment, averaged across all 1,121 cases. Blurring the most salient pixels of an input clip results in an immediate and substantial drop in average prediction confidence, while blurring random pixels shows a much less substantial drop in confidence. This indicates that the model relies on the most salient pixels according to the Grad-CAM saliency maps to make its prediction, supporting the saliency maps' use in visual analysis.

Of the 20 clips analyzed, in all clips, the most salient areas did not shift or deform significantly over time, and the frames with the highest saliency pixels always appeared between the release of the ball and the moment the ball falls to the ground.

\begin{figure}[htbp]
    \centering
    \includegraphics[width=\linewidth]
    {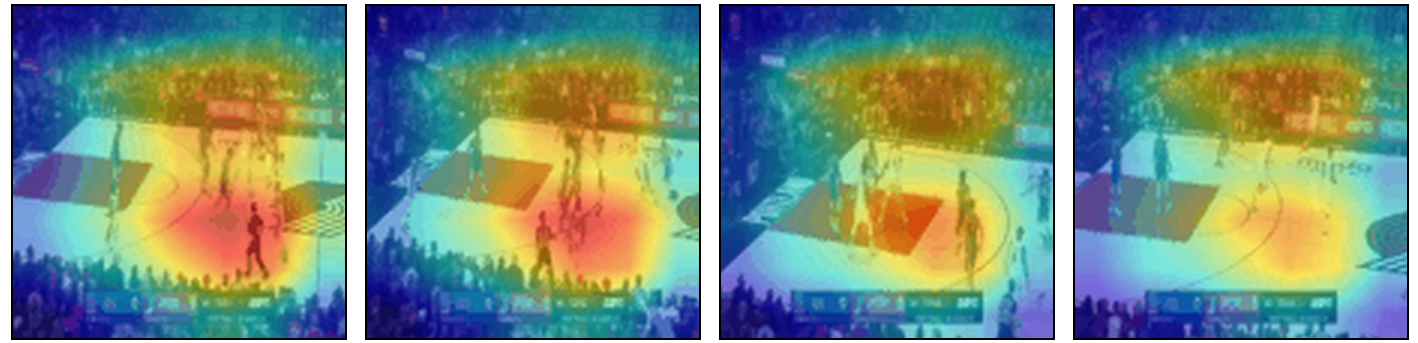}
    \caption{Representative Grad-CAM sequence. Frames progress left to right. Warmer colors indicate higher saliency.}
    \label{fig:gradcam}
\end{figure}

In 5/5 2-point clips, the shooter and net were most salient. In 5/5 mid-range and 5/5 free-throw clips, the area under the net was most salient. In 2/5 of 3-point shots, the shooter was salient, while in the other 3/5 the shooter and courtside crowd were most salient, though the crowd showed less saliency than the player. Figure~\ref{fig:gradcam} shows an example of this shooter-crowd saliency. These observations suggest that the model generally relies on central frame regions containing the shooter and net, while the specific emphasis varies across shot categories. However, the lower-intensity crowd saliency found in 3-point shots indicates that broadcast features may influence predictions, so the visual analysis cannot completely rule out shortcut learning. The centre (60$\times$60) pixel region of the test clips was blurred for the targeted set. The control set applied the same blurring but in random locations. Although central blurring does not definitively separate between player, ball, net, or background, it provides an estimate for this preliminary audit. Compared to the baseline accuracy of 69.3\%, model accuracy dropped to 50.9\% on the random blur dataset (a decrease of 18.4 percentage points) and further dropped to 43.7\% on the targeted blur dataset (a decrease of 25.6 percentage points). While the model's accuracy drops in both datasets, the targeted blur dataset results in an additional 7.2-percentage-point decrease. This suggests that the model relies on the targeted region to make accurate predictions.

\section{Discussion and Future Work}

The applied study in this paper provides important new insights.
Our work bridges computer vision, XAI and sports analytics. We offer:
i) explanation workflows adaptable to other CNN-based action recognition models;
ii) possible advances to sports analytics by auditing shortcut learning in models to reduce unfair penalization;
iii) a role for XAI in any spatiotemporal setting to identify where learning methods have weaknesses, towards improved designs.

Future work will compare shortcut learning across architectures. For example, vision transformers may rely on spurious correlations such as the crowd to a lesser extent. Further, deep learning models can be trained and tested with shortcut mitigation efforts such as masking out the scoreboard, crowd, and other potential spurious correlations.

\bibliography{references}

\end{document}